\documentclass[letterpaper, 10 pt, conference]{ieeeconf}  % Comment this line out if you need a4paper

\IEEEoverridecommandlockouts                              % This command is only needed if 
\usepackage{graphics} % for pdf, bitmapped graphics files
\usepackage{epsfig} % for postscript graphics files
\usepackage{mathptmx} % assumes new font selection scheme installed
\usepackage{times} % assumes new font selection scheme installed
\usepackage{amsmath} % assumes amsmath package installed
\usepackage{amssymb}  % assumes amsmath package installed

\title{\LARGE \bf
Conceptual Cognitive Maps Formation with \\ Neural Successor Networks and Word Embeddings
}

\author{Paul Stöwer$^{1}$, Achim Schilling$^{2}$, Andreas Maier$^{3}$, and Patrick Krauss.$^{4}$% <-this % stops a space
\thanks{$^{1}$Paul Stöwer is with the Pattern Recognition Lab, University Erlangen-Nürnberg, Erlangen, Germany
        {\tt\small paul.stoewer@fau.de}}%
\thanks{$^{2}$Achim Schilling is with the Neuroscience Lab, University Hospital Erlangen, Erlangen, Germany
        {\tt\small achim.schilling@fau.de}}%
\thanks{$^{3}$Andreas Maier s with the Pattern Recognition Lab, University Erlangen-Nürnberg, Erlangen, Germany
        {\tt\small andreas.maier@fau.de}}%
\thanks{$^{4}$Patrick Krauss is with the Neuroscience Lab, University Hospital Erlangen, Erlangen, Germany
        {\tt\small patrick.Krauss@fau.de}}%
}

\begin{document}

\maketitle
\thispagestyle{empty}
\pagestyle{empty}

%%%%%%%%%%%%%%%%%%%%%%%%%%%%%%%%%%%%%%%%%%%%%%%%%%%%%%%%%%%%%%%%%%%%%%%%%%%%%%%%
\begin{abstract}

The human brain possesses the extraordinary capability to contextualize the information it receives from our environment. The entorhinal-hippocampal plays a critical role in this function, as it is deeply engaged in memory processing and constructing cognitive maps using place and grid cells. Comprehending and leveraging this ability could significantly augment the field of artificial intelligence. The multi-scale successor representation serves as a good model for the functionality of place and grid cells and has already shown promise in this role. Here, we introduce a model that employs successor representations and neural networks, along with word embedding vectors, to construct a cognitive map of three separate concepts. The network adeptly learns two different scaled maps and situates new information in proximity to related pre-existing representations. The dispersion of information across the cognitive map varies according to its scale - either being heavily concentrated, resulting in the formation of the three concepts, or spread evenly throughout the map. We suggest that our model could potentially improve current AI models by providing multi-modal context information to any input, based on a similarity metric for the input and pre-existing knowledge representations.

\end{abstract}

%%%%%%%%%%%%%%%%%%%%%%%%%%%%%%%%%%%%%%%%%%%%%%%%%%%%%%%%%%%%%%%%%%%%%%%%%%%%%%%%
\section{INTRODUCTION}

The memories in our brains make up our past experience and shape how we see the world. The hippocampus is heavily involved in the domain of memory processing and transfers short term to long term memories \cite{kryukov_role_2008}\cite{reddy2021human}\cite{tulving1998episodic}. Furthermore a main function of the hippocampus is navigation in both spatial and non-spatial abstract mental spaces \cite{epstein2017cognitive}\cite{killian_grid_2018}\cite{okeefe_hippocampus_1971}\cite{park2021inferences}.

\begin{figure}[ht]
\begin{center}
\includegraphics[width=9cm]{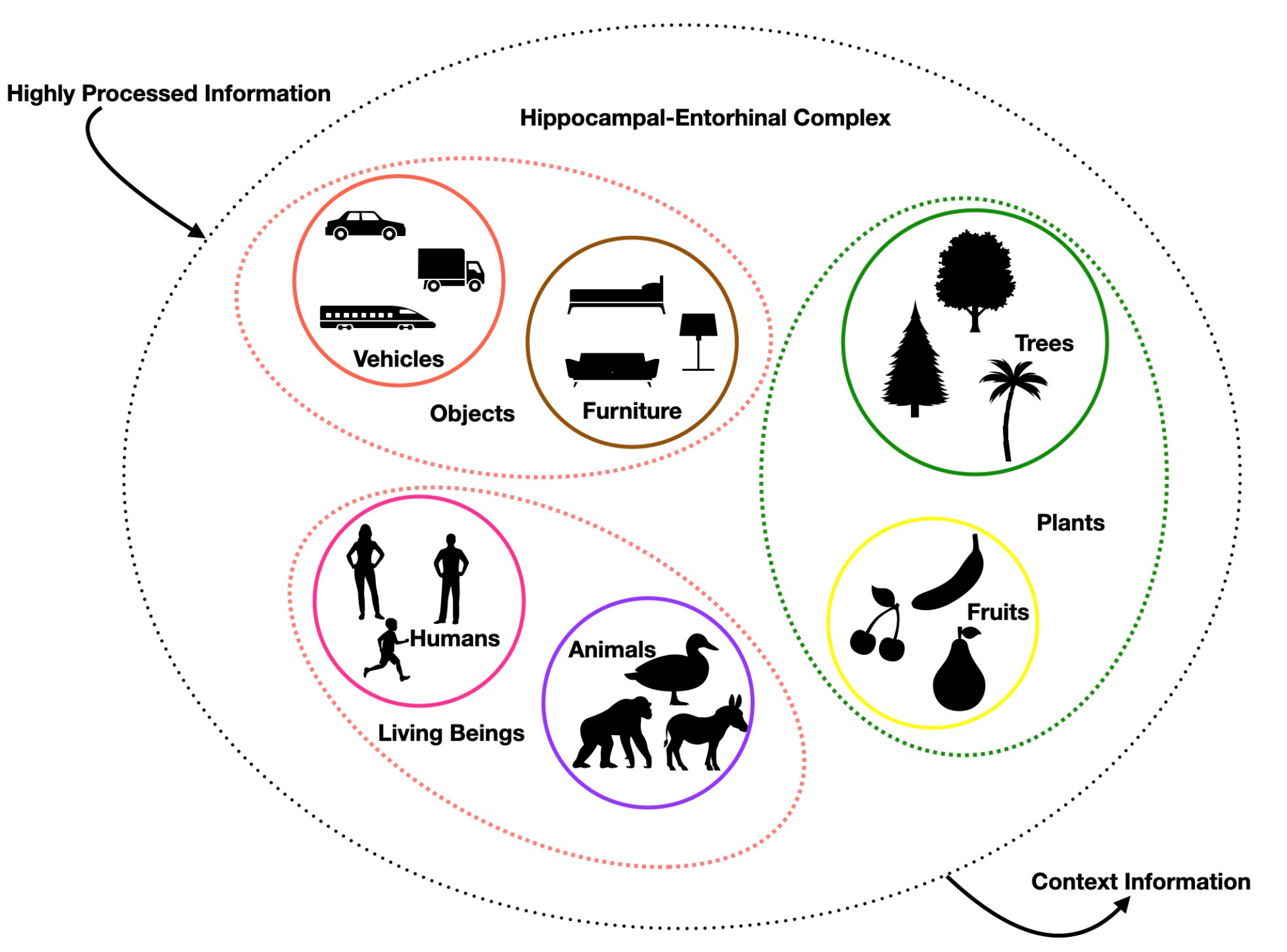}
\end{center}
\caption{The entorhinal-hippocampal complex receives highly processed information from other brain regions. 
The entorhinal-hippocampal complex is expected to build cognitive maps via place and grid cells with these information. Cognitive Maps are meant to display arbitrary information and their relationship to each other. The varying grid size of the grid cells in the entorhinal complex enable additionally a scale which can range from specific information to wider concepts, depending on the inputs received and the outputs sought.} 
\label{gamma}
\end{figure}

The hippocampus is expected to build cognitive maps, which can display arbitrary information and their relationship to each other \cite{o1978hippocampus}. Place and grid cells are cell types in the entorhinal-hippocampal complex, which are involved in the formation of these maps \cite{moser2017spatial}. 

In addition, memory can be represented at varying degrees of detail along the longitudinal axis of the hippocampus, such as in diverse spatial resolutions \cite{collin2015memory}. These varying scales assist in navigation over differing horizons in terms of spatial navigation \cite{brunec2019predictive}. When we consider abstract conceptual spaces, these diverse scales may denote varying levels of abstraction \cite{milivojevic2013mnemonic}. Broadly speaking, these multi-scale cognitive maps facilitate flexible planning, enable the broad generalization of concepts, and foster intricate representation of information \cite{momennejad2020learning}.

Inspired by the properties of the entorhinal-hippocampal complex, we use the successor representation (SR). The SR is a mathematical model, which can be used to model place cell activity \cite{stachenfeld2017hippocampus}. Furthermore, models for multi-scale successor representations are proposed to enable maps with different scales \cite{MSSR} and the SR can also be used for flexible sequence generation \cite{mcnamee2021flexible}.

In previous studies, we could already demonstrate that we can use the SR and artificial neural networks to model cognitive maps for different scenarios. We have recreated place cell fire patterns in spatial navigation experiments with rodents and have built a simplified language model \cite{stoewer_neural_2022}. Additionally, we built cognitive maps with handcrafted features of different animal species and used the map to interpolate novel and hidden information \cite{stoewer2023neural}. Furthermore, we showed that word class representations spontaneously emerge in a deep neural network trained on next word prediction \cite{surendra2023word}.
Here, we go a step further and use generative features of large language models to build a cognitive map. In particular, the word embeddings of these models are used as semantic features for different categories of objects.
This might serve as a first building block, to incorporate and access further information related to words and sentences, which could lead to the formation of concepts using the properties of navigation and memory processing inspired by the hippocampus. 

\section{Methods}

\subsection{Successor Representation}

\begin{equation}\label{successor_representation}
\centering
V(s) = E[\sum^{\infty}_{t=0}\gamma^t R(s_t)|s_0=s] 
\end{equation}

The Successor Representation was designed to describe the potential future reward $V(s)$ from a current state $s$ over a time period $t$ counting all rewards $R(s_t)$ to all successor states. The discount factor $\gamma$ influences the weight of the successor states to the reward $V(s)$ \cite{SR_Original} (cf. eq.\ref{successor_representation}). Stachenfeld et. al used the SR to successfully model place cell behaviour \cite{stachenfeld2017hippocampus}.

\begin{equation}\label{SR_refactor}
    V(s) = \sum_{s'} M(s,s')R(s') \hspace{1cm} M = \sum^{\infty}_{t=0}\gamma^t T^t
\end{equation}

Cognitive maps can be constructed with the transition probability matrix of the state space, which gives the relationship between the states. The matrix can then also be used to calculate the successor representation matrix (cf. eq.\ref{SR_refactor}).

\begin{figure}[ht]
\begin{center}
\includegraphics[width=7cm]{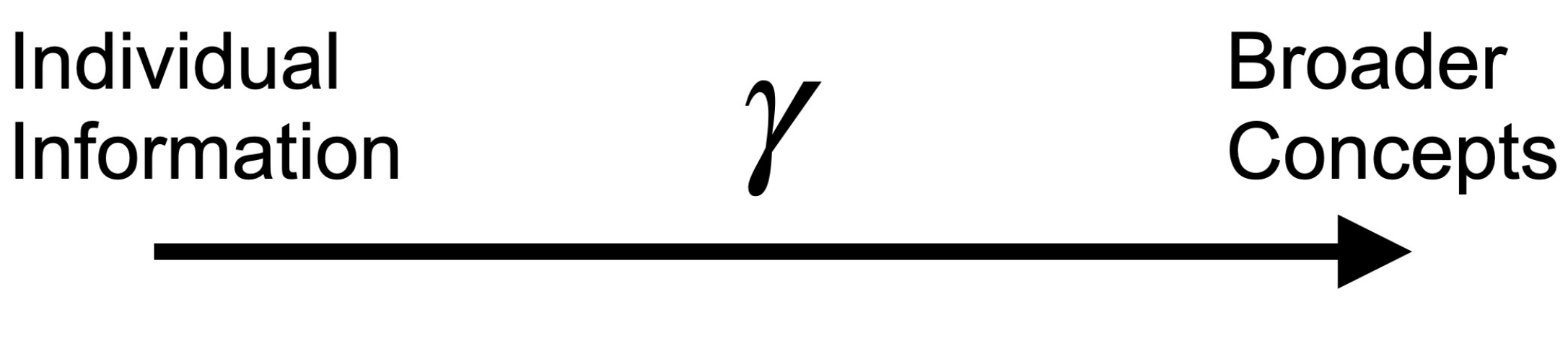}
\end{center}
\caption{The discount factor $\gamma$ might describe the varying grind size in the entorhinal cortex. The different scales could be used to create different degrees of abstraction in the cognitive maps and encode information from an individual level to  broader concepts.} 
\label{gamma}
\end{figure}

The discount factor $\gamma$ might furthermore be used to model the varying grid size in the entorhinal cortex \cite{stachenfeld2017hippocampus} and therefore be a tool to encode information from broader concepts to more detailed individual properties \cite{milivojevic2013mnemonic}(cf. Figure \ref{gamma}).

\subsection{Word Category Data Set}

The set up cognitive map is based on word embeddings of 20 words of 3 categories: Animals, vehicles and furniture. 10 different words for each category are used as validation data.
The word embeddings hold features of each word and can be used to compare words an their similarity to each other. We used the spacy library to calculate the embeddings.
The transition probabilities of the state space where calculated via the similarity function of spacy, which uses the cosine similarity between the embeddings ($A_s$ and $B_s'$ (cf. eq. \ref{data_set_transitions})).

\begin{equation}\label{data_set_transitions}
\centering
T(s,s')=cos(\theta)=\frac{A_s\cdot B_{s'}}{||A_s||||B_{s'}||}
\end{equation}

\subsection{Neural Network Architecture}

\begin{figure}[ht]
\begin{center}
\includegraphics[width=9cm]{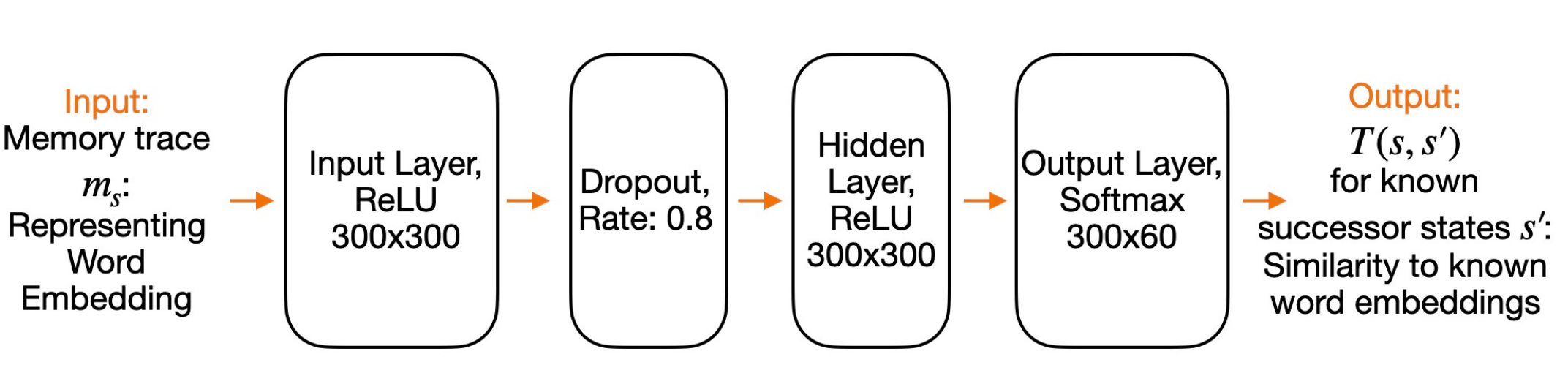}
\end{center}
\caption{The architecture of the used model consists of 4 layers. The word embeddings serve as input, a dropout layer helps with the adaptability and the output layer points to the 60 saved "memories" of the training word embeddings} 
\label{architecture}
\end{figure}

In order to learn the successor representation of the state space, we set up a neural network. The network receives as input the vector of the word embedding. Right after the input layer, a Dropout layer with a rate of $0.8$ is placed, to increase the robustness of novel unseen inputs. One hidden layer is used, before the output layer points with a Softmax activation function to the 60 known states, which represent the word embeddings of the training data (cf. Figure \ref{architecture}). Two different networks are trained for two different $\gamma$ with $[1.0\&0.7]$ and $t=5$. The training was performed over 500 epochs, a batch size of 20 and with a learning rate of $1e-5$.

\subsection{Multi-dimensional scaling}

We used multi-dimensional scaling (MDS) to project the data on to a 2D surface, although T-distributed stochastic neighbor embedding (t-SNE) is a common method for producing low-dimensional embeddings from high-dimensional data \cite{van2008visualizing}. Nevertheless, t-SNE's low-dimensional projections can be highly sensitive to specific parameter settings \cite{wattenberg2016use}, are susceptible to noise, and may often scramble rather than retain the global structure in data \cite{vallejos2019exploring, moon2019visualizing}.

In contrast, multi-Dimensional-Scaling (MDS) offers an effective method for visualizing high-dimensional point clouds by projecting them onto a 2-dimensional plane \cite{torgerson1952multidimensional, kruskal1964nonmetric,kruskal1978multidimensional,cox2008multidimensional}. A key benefit of MDS is that it does not require parameter tuning, and it maintains all mutual distances between the points, thus preserving both the global and local structure of the data it represents. 

When interpreting patterns as points in high-dimensional space and dissimilarities between patterns as distances between corresponding points, MDS is an elegant method to visualize high-dimensional data. By color-coding each projected data point of a data set according to its label, the representation of the data can be visualized as a set of point clusters. For instance, MDS has already been applied to visualize for instance word class distributions of different linguistic corpora \cite{schilling2021analysis}, hidden layer representations (embeddings) of artificial neural networks \cite{schilling2021quantifying,krauss2021analysis}, structure and dynamics of highly recurrent neural networks \cite{krauss2019analysis, krauss2019recurrence, krauss2019weight, metzner2023quantifying}, or brain activity patterns assessed during e.g. pure tone or speech perception \cite{krauss2018statistical,schilling2021analysis}, or even during sleep \cite{krauss2018analysis,traxdorf2019microstructure,metzner2022classification,metzner2023extracting}. 

In all these cases the apparent compactness and mutual overlap of the point clusters permits a qualitative assessment of how well the different classes separate.

\subsection{Generalized Discrimination Value (GDV)}

We used the GDV to calculate cluster separability as published and explained in detail in \cite{schilling2021quantifying}. Briefly, we consider $N$ points $\mathbf{x_{n=1..N}}=(x_{n,1},\cdots,x_{n,D})$, distributed within $D$-dimensional space. A label $l_n$ assigns each point to one of $L$ distinct classes $C_{l=1..L}$. In order to become invariant against scaling and translation, each dimension is separately z-scored and, for later convenience, multiplied with $\frac{1}{2}$:
\begin{align}
s_{n,d}=\frac{1}{2}\cdot\frac{x_{n,d}-\mu_d}{\sigma_d}.
\end{align}
Here, $\mu_d=\frac{1}{N}\sum_{n=1}^{N}x_{n,d}\;$ denotes the mean, and $\sigma_d=\sqrt{\frac{1}{N}\sum_{n=1}^{N}(x_{n,d}-\mu_d)^2}$ the standard deviation of dimension $d$.
Based on the re-scaled data points $\mathbf{s_n}=(s_{n,1},\cdots,s_{n,D})$, we calculate the {\em mean intra-class distances} for each class $C_l$ 
\begin{align}
\bar{d}(C_l)=\frac{2}{N_l (N_l\!-\!1)}\sum_{i=1}^{N_l-1}\sum_{j=i+1}^{N_l}{d(\textbf{s}_{i}^{(l)},\textbf{s}_{j}^{(l)})},
\end{align}
and the {\em mean inter-class distances} for each pair of classes $C_l$ and $C_m$
\begin{align}
\bar{d}(C_l,C_m)=\frac{1}{N_l  N_m}\sum_{i=1}^{N_l}\sum_{j=1}^{N_m}{d(\textbf{s}_{i}^{(l)},\textbf{s}_{j}^{(m)})}.
\end{align}
Here, $N_k$ is the number of points in class $k$, and $\textbf{s}_{i}^{(k)}$ is the $i^{th}$ point of class $k$.
The quantity $d(\textbf{a},\textbf{b})$ is the euclidean distance between $\textbf{a}$ and $\textbf{b}$. Finally, the Generalized Discrimination Value (GDV) is calculated from the mean intra-class and inter-class distances  as follows:
\begin{align}
\mbox{GDV}=\frac{1}{\sqrt{D}}\left[\frac{1}{L}\sum_{l=1}^L{\bar{d}(C_l)}\;-\;\frac{2}{L(L\!-\!1)}\sum_{l=1}^{L-1}\sum_{m=l+1}^{L}\bar{d}(C_l,C_m)\right]
 \label{GDVEq}
\end{align}

\noindent whereas the factor $\frac{1}{\sqrt{D}}$ is introduced for dimensionality invariance of the GDV with $D$ as the number of dimensions.

\vspace{0.2cm}\noindent Note that the GDV is invariant with respect to a global scaling or shifting of the data (due to the z-scoring), and also invariant with respect to a permutation of the components in the $N$-dimensional data vectors (because the euclidean distance measure has this symmetry). The GDV is zero for completely overlapping, non-separated clusters, and it becomes more negative as the separation increases. A GDV of -1 signifies already a very strong separation.

\section{RESULTS}

\begin{figure}[ht]
\begin{center}
\includegraphics[width=9cm]{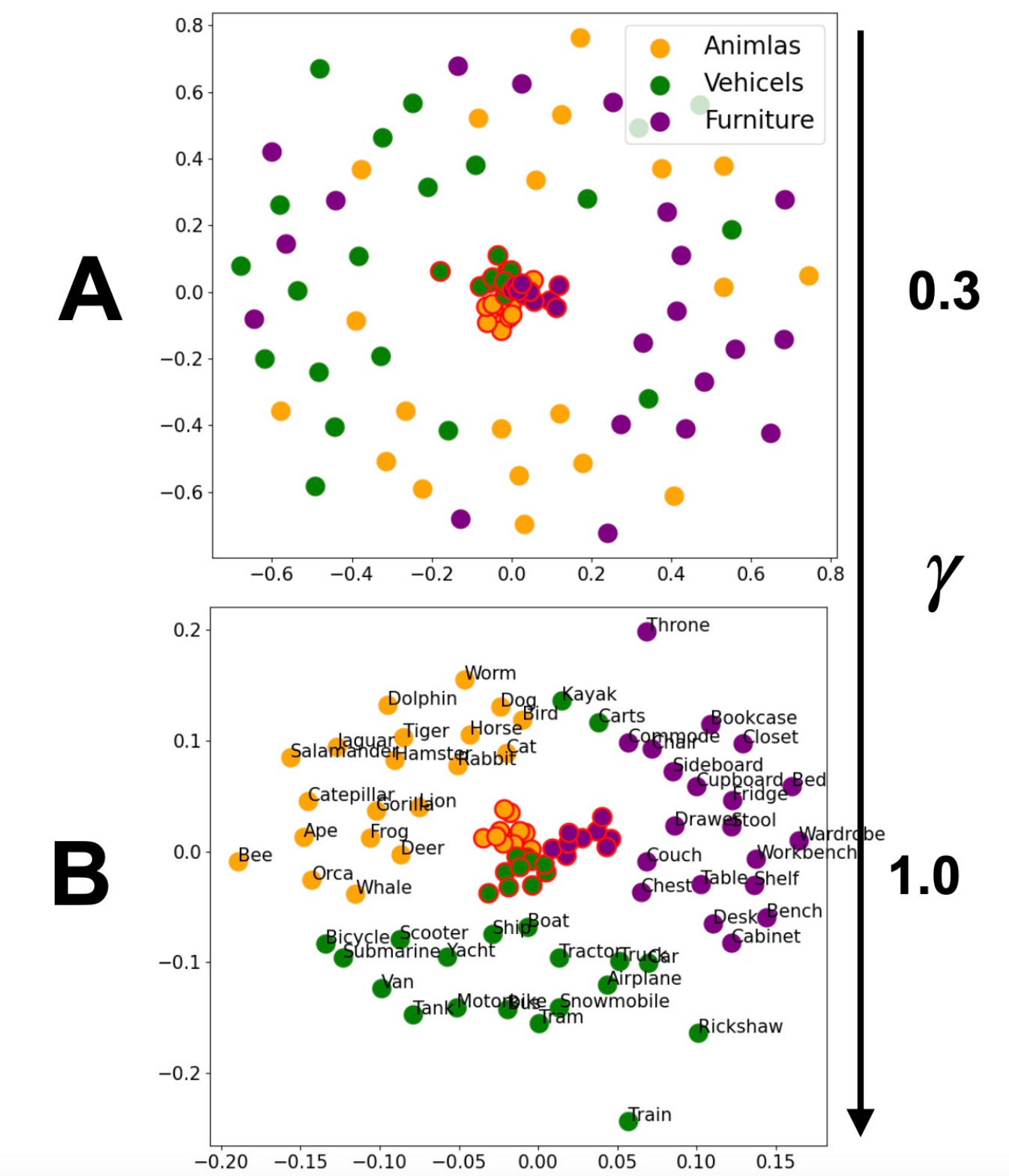}
\end{center}
\caption{Two MDS projections for the prediction of all training and validation (red encircled) word embeddings can be seen. \textbf{A:} In the mapping of the SR with $\gamma=0.3$ we see that the predictions of the training data is spread evenly over the feature space. The validation samples form overlapping clusters. \textbf{B:} The MDS for the SR with $\gamma=1.0$ shows cluster formation for the training and validation data. The validation cluster are however closer to each other in comparison to the training data. } 
\label{MDS}
\end{figure}

The neural network is able to learn the set up cognitive maps successfully. We projected the predictions of the network for all training and validation word embedding vectors using multi-dimensional scaling (MDS). Remarkably, three object category representations spontaneously emerge in the network with a discount factor $\gamma=1.0$ (cf. Figure \ref{MDS} B). In comparison the discount factor $\gamma=0.3$ leads to an evenly spread feature space (cf. Figure \ref{MDS} A). Furthermore, we calculated the generalized discrimination value (GDV) \cite{schilling2021quantifying} to quantify the clustering of the representations. The GDV for $\gamma=1.0$ considering all data points is $-0.44$, indicating strong clustering. If we only consider the training samples, the GDV is $-0.43$ and for the validation samples $-0.39$. However, the predictions derived from the network with the lower discount factor $\gamma=0.3$ do not form dense clusters. This is also reflected by respective GDVs, indicating weak clustering. Here, for all data points the GDV is $-0.38$, whereas the GDV for the training samples is even smaller with $-0.35$ and for the validation data with $-0.32$.

\section{DISCUSSION}

Our study demonstrates that it is possible to construct and learn differently scaled cognitive maps by using similarity measures between inputs and known states reflecting word embeddings. We propose that this mechanism can be paralleled with the process of memory recognition, where novel sensory inputs are compared to previously stored memories. Hence, we can extract contextual information from new inputs using the cognitive maps created by the learned memories. This strategy could potentially enhance the performance of current AI systems.

\begin{figure}[ht]
\begin{center}
\includegraphics[width=8cm]{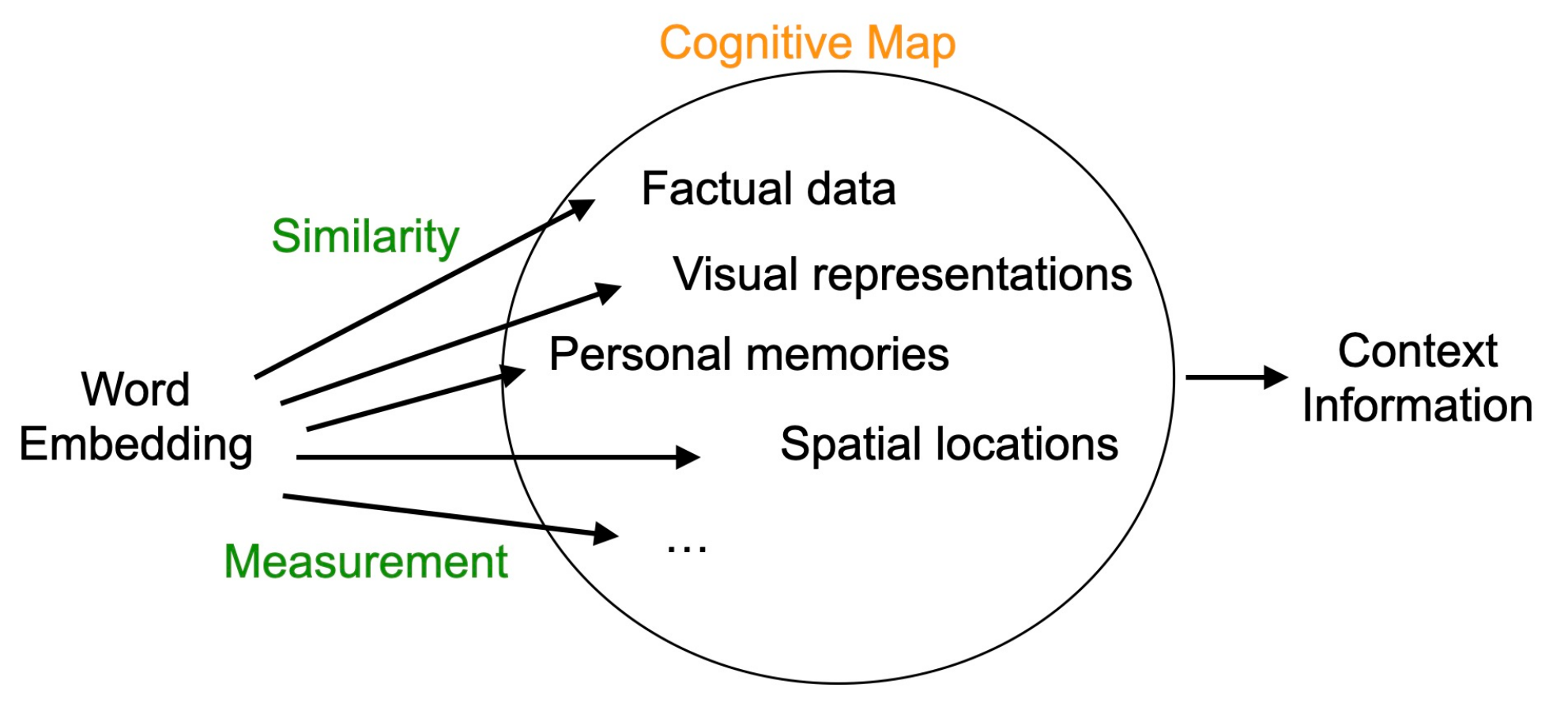}
\end{center}
\caption{The schematic sketch shows how a language model can profit from cognitive maps by using word embeddings. A similarity measure for the input word or phrase to past memories, where the input was also present, can give more context information for the novel input. Therefore we could retrieve multi modal information like visual representation, spatial locations or personal/emotional memories to the according inputs and gain context information on the present task.} 
\label{CognitiveMap}
\end{figure}

For instance, let's consider large language models (LLMs). Over the past year, there has been significant improvement in LLMs, and they are particularly proficient at text generation tasks. Nonetheless, despite their superior formal competence, they still fall short in terms of functional competence \cite{mahowald2023}. Our model could potentially overcome these limitations. A similarity measure of word embeddings to arbitrary additional information, like factual data, visual representations or spatial locations could provide more context information to text input (cf. Figure \ref{CognitiveMap}). Furthermore, it could  use this technique to "fact check" its proposition with a memory database, which is not only generated by predictions from the training data. How our proposed model can be incorporated into a LLM will be part of future research.
In general, the model could be also useful for many other applications, to contribute more context awareness depending on the current situation.

\section*{ACKNOWLEDGMENT}

This work was funded by the Deutsche Forschungsgemeinschaft (DFG, German Research Foundation): grants KR\,5148/2-1 (project number 436456810), KR\,5148/3-1 (project number 510395418) and GRK\,2839 (project number 468527017) to PK, and grant SCHI\,1482/3-1 (project number 451810794) to AS, and by the Emerging Talents Initiative (ETI) of the University Erlangen-Nuremberg (grant 2019/2-Phil-01 to PK).

\bibliographystyle{unsrt}
\bibliography{literature}

\end{document}